\documentclass[runningheads,a4paper]{llncs}

\usepackage{amssymb}
\usepackage{graphicx}
\usepackage{color}
\usepackage{subfig}

\usepackage{url}
\newcommand{\keywords}[1]{\par\addvspace\baselineskip
\noindent\keywordname\enspace\ignorespaces#1}

\newcommand*\chains{\includegraphics{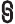}}
\usepackage{array}
\newcolumntype{L}[1]{>{\raggedright\let\newline\\\arraybackslash\hspace{0pt}}m{#1}}
\newcolumntype{C}[1]{>{\centering\let\newline\\\arraybackslash\hspace{0pt}}m{#1}}
\newcolumntype{R}[1]{>{\raggedleft\let\newline\\\arraybackslash\hspace{0pt}}m{#1}}

\begin{document}

\mainmatter  % start of an individual contribution

% first the title is needed
\title{Direct Prediction of Cardiovascular Risk in Chest CT using Real-Time ConvNet Regression}

% a short form should be given in case it is too long for the running head
\titlerunning{Direct and Real-Time Cardiovascular Risk Prediction}

% the name(s) of the author(s) follow(s) next
%
% NB: Chinese authors should write their first names(s) in front of
% their surnames. This ensures that the names appear correctly in
% the running heads and the author index.
%

\author{Bob D. de Vos\inst{1}
	\and
	Nikolas Lessmann\inst{1}
	\and
	Pim A. de Jong\inst{2}
	\and\\
	Max A. Viergever\inst{1}
	\and
	Ivana I\v{s}gum\inst{1}}
%
%\author{Anonymous
%	\and
%	Anonymous
%	\and
%	Anonymous
%	\and\\
%	Anonymous
%	\and
%	Anonymous}

%
\authorrunning{Bob D. de Vos et al.}
% (feature abused for this document to repeat the title also on left hand pages)

% the affiliations are given next; don't give your e-mail address
% unless you accept that it will be published
\institute{Image Sciences Institute, University Medical Center Utrecht, the Netherlands 
			\and
			Department of Radiology, University Medical Center Utrecht, the Netherlands}

%\institute{Anonymous
%	\and
%			Anonymous}

%toctitle{Lecture Notes in Computer Science}
%\tocauthor{Authors' Instructions}
\maketitle

\begin{abstract}
Coronary artery calcium (CAC) burden quantified in low-dose chest CT is a predictor of cardiovascular events. We propose an automatic method for CAC quantification, circumventing intermediate segmentation of CAC. The method determines a bounding box around the heart using a ConvNet for localization. Subsequently, a dedicated ConvNet analyzes axial slices within the bounding boxes to determine CAC quantity by regression.
A dataset of 1,546 baseline CT scans was used from the National Lung Screening Trial with manually identified CAC. The method achieved an ICC of 0.98 between manual reference and automatically obtained Agatston scores. Stratification of subjects into five cardiovascular risk categories resulted in an accuracy of 85\% and Cohen's linearly weighted $\kappa$ of 0.90. The results demonstrate that real-time quantification of CAC burden in chest CT without the need for segmentation of CAC is possible.

\end{abstract}

\section{Introduction}
Cardiovascular disease (CVD) is the leading cause of death globally. Coronary artery calcium (CAC) indicates presence of CVD, and CAC burden has been shown to be a strong and independent predictor of cardiovascular events (CVEs) \cite{Yebo12}. Quantifying CAC (calcium scoring) is standardly performed in dedicated ECG-triggered cardiac CT scans without contrast enhancement. However, calcium scoring can also be performed in non-dedicated scans visualizing the heart, e.g. chest CT scans. CAC burden determined in low-dose chest CT scans acquired in lung cancer screening studies has also been shown to predict all-cause mortality \cite{Jaco10}. In addition, the two largest lung screening trials, the National Lung Screening Trial (NLST) and the Dutch-Belgian lung cancer screening trial (NELSON),  demonstrated that not lung cancer but CVD is the leading cause of mortality in populations of heavy smokers \cite{Chiles15,Jaco10,Klav09}. Hence, determining calcium score as an unrequested finding in lung cancer screening scans has recently gained attention.

In clinical routine, coronary artery calcifications are scored manually. This requires an expert to identify CAC lesions. Especially in non-dedicated chest CT scans this is a cumbersome process. Lack of ECG-triggering results in artifacts that are caused by cardiac motion, thus affecting appearance of coronary calcifications. Low irradiation dose results in high noise levels, which makes it hard to identify small calcifications. In addition, subjects undergoing screening are at high risk of CVD due to history of heavy smoking, thus often present many calcified lesions requiring substantial manual interaction in scoring. %Moreover, chest scans are larger than cardiac CT scans, hence, in manual scoring, large number of slices needs to be inspected.

To allow fast identification of subjects at risk of CVEs in lung cancer screening settings at low cost, qualitative instead of quantitative scoring has been proposed categorizing subjects into none, mild, moderate, or heavy risk groups \cite{Chiles15}. Even though simple and fast, such analyses requires experts to inspect the scans. To overcome this limitation, several automatic calcium scoring methods for chest CT have been proposed \cite{Isgu12,Less16,Xie14}. These methods first select high density voxels that are assumed to contain calcium with the clinically accepted threshold of 130\,HU. The selected voxels are than classified by supervised machine learning methods to separate CAC from other calcifications. I\v{s}gum et al. \cite{Isgu12} proposed a method that used a map providing a priori probabilities for the spatial appearance of CAC. Thereafter, a two-stage classification employing nearest neighbor and support vector classifiers was designed to identify coronary calcifications using texture, size, and spatial features that were derived from the spatial a priory probability map. Xie et al. \cite{Xie14} identified CAC lesions as connected groups of high-intensity voxels in a roughly segmented heart that was found by segmentations of the lungs, bone, aorta and fatty tissue. Lessmann et al. \cite{Less16} identified coronary artery calcifications using a convolutional neural network (ConvNet) that analyzed high density voxels in a bounding box around the heart.

We propose a method that determines calcium score directly from the CT scan that circumvents intermediate segmentation of coronary calcifications. The method first localizes the heart using a ConvNet for classification and subsequently quantifies CAC burden in axial image slices directly using a ConvNet for regression. The method allows direct and real-time quantification of cardiovascular risk in lung cancer screening participants.

\section{Data}
A subset of 1,546 chest CT scans (911 men, age range: 54--74) was randomly selected from a set of 6,000 available baseline scans from the NLST. Scans were selected with 1:2 ratio of deaths:survivors to match the overall ratio in our data set. All scans were acquired during inspiratory breath-hold without contrast enhancement. Scans were acquired in 31 different hospitals with 120 or 140\,kVp tube voltage and 30--160\,mAs tube current. Axial images slices were reconstructed with varying kernels, varying thickness (1.00--3.00\,mm), varying increments (0.63--3.00\,mm), and with varying in-plane resolutions (0.49--0.98\,mm per voxel). In our study, scans with less than 100 slices or slices thicker than 3\,mm were not considered, because they were not adequate for calcium scoring. Furthermore, the scans were resampled to 3.00\,mm slice thickness and 1.50\,mm slice increment to make the scans suitable for calcium scoring \cite{Rutt10}.

The reference standard was defined by one of three observers who manually identified CAC lesions in the scans. Following the clinical procedure, CAC lesions were defined as 26-connected components containing voxels above 130\,HU threshold with a minimum size of 1.5\,mm$^3$. For each subject Agatston and volume calcium scores were computed. These scores were used as reference standard for development, training, and evaluation of the method. Based on the Agatston score, each subject was assigned to one of five cardiovascular risk groups: very low ($<$1), low (1--10), moderate (10--100), moderately high (100--400), and high ($\geq$400).
%Because the images acquired in lung cancer screening do not allow precise calcium quantification \cite{Jaco10a}, participants were categorized in five risk groups: low ($<$100), moderate (100-400) and high ($\ge$400).

\section{Method}
%Our algorithm quantifies coronary artery calcification (CAC) in axial slices of the heart. Unlike previous methods, which first classified CAC and quantified it afterwards, CAC quantity is determined directly from an input image slice by using regression with a ConvNet.

We propose a ConvNet design to directly quantify CAC from axial image slices using regression. This mimics clinical analysis, where CAC burden is quantified in axial slices with the Agatston score \cite{Agat90} for CVD risk categorization \cite{Detr08,Agat90}. The score is determined as follows, in each slice the area of each CAC lesion is multiplied by a weight factor that is defined by the maximum CT number in the lesion (1: 130--200\,HU, 2: 200--300\,HU, 3: 300--400\,HU, 4: $\geq$400\,HU). The total patient Agatston score is obtained by addition of CAC lesion scores in all slices. Alternatively, CAC can be quantified with the CAC volume score (i.e. the total volume of all CAC lesions in scan). The volume score has been shown to be more reproducible and -- although without clinical implications -- it is often calculated. 

To facilitate analysis within the heart region only, a bounding box around the heart is first extracted using a method described by De Vos et al. \cite{de_vos16}. This method uses three independent ConvNets that each determine presence of the heart in axial, coronal, or sagittal image slices. The combination of these per-slice probabilities yields a 3D bounding box around the heart.

\begin{figure}[h]
	\centering
	\includegraphics[width=.7\textwidth]{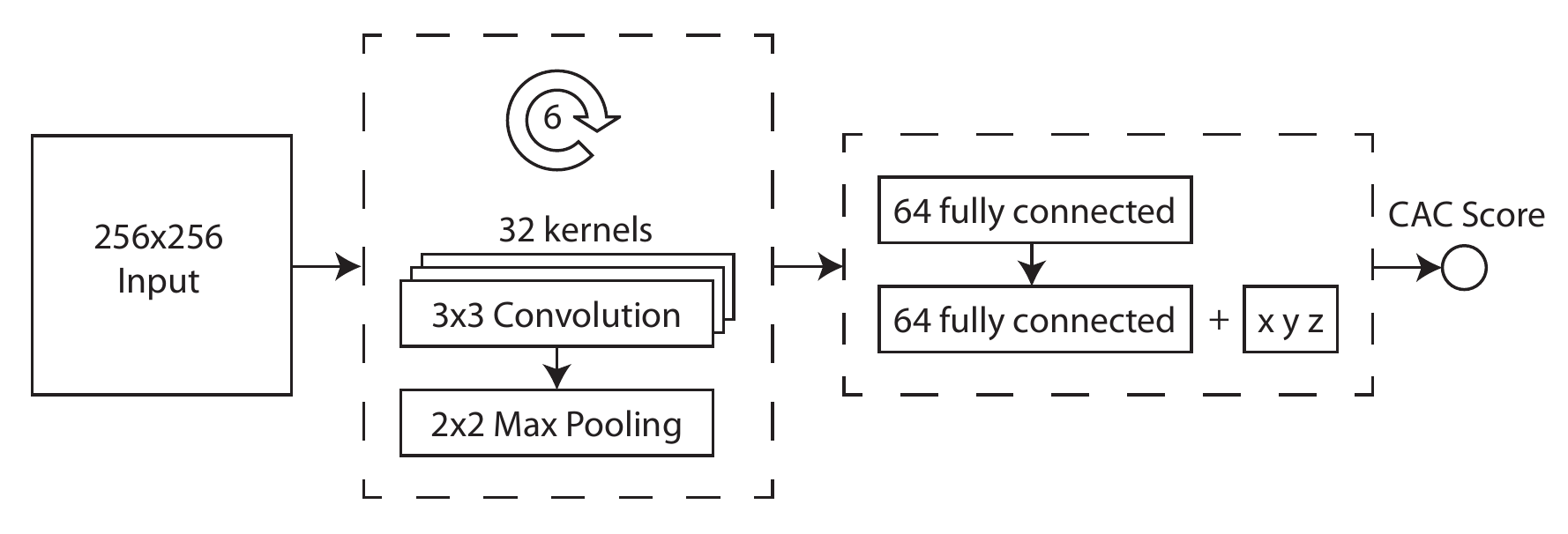}
	\caption{The architecture of the regression ConvNet that was used for CAC quantification. It consists of six convolutional layers, two fully connected layers, and has a single output node. Batch normalization is applied in each hidden layer and exponential linear units are used for activation. The network expects image slices of $256\times256$~pixels input and outputs a single continuous calcium score.}
	\label{fig:network}
\end{figure}

Axial image slices cropped to the cardiac bounding box are used as input for a dedicated regression ConvNet (Figure~\ref{fig:network}). The ConvNet contains six layers of 3x3 convolution kernels with 2x2 max pooling, two fully connected layers, and thereafter a single output node. Batch normalization is applied after each layer with exponential linear units used for activation. To enable correct computation of area or volume of calcified lesions, voxel dimensions are appended to the first connected layer. The regression ConvNet is trained to output either an Agatston or CAC-volume score.

%Because the Agatston is originally used in cardiac CT of 3\,mm slice thickness and increment, the score was corrected for 

\section{Experiments}
To train the ConvNet, 1,158 images were randomly selected. Among them 772 were used to design and 386 to validate the network. An independent set of 388 images was selected to test the method and was not used during method development in any way. 

Input image slices were first cropped to the cardiac bounding box and were thereafter 0-padded to $256\times256$~pixels to obtain a fixed input size for the regression ConvNet.

%(The image slices could've been too large because the heart was large or localization could've failed. I need to find out why.)
The regression ConvNet was trained in 50 epochs with mini-batches of 100 randomly ordered image slices. After every epoch the performance was evaluated on the validation set and the optimal network was chosen based on the training and the validation errors, which were defined as the mean of absolute differences between target and predicted values. Adam \cite{king_14} was used as optimizer with a learning rate of~$0.01$, a first moment exponential decay rate of~0.9, and a second moment exponential decay rate of~0.999. 

Four experiments were performed for both Agatston and volume scoring. In the first experiment a regression ConvNet was trained with image slices as input and the per-slice manually determined CAC score as the target regression value. The ConvNet was trained with separate kernels per convolutional layer. Thereafter, the same setup was used but with shared kernels (shared weights but separate biases per kernel) to limit the number of parameters. In the third experiments the same network design was trained, but with log transformed target regression value ($y_t = ln(y + 1)$, where $y_t$ is the transformed target value and $y$ the reference CAC score. In the final experiment, the same log transformed target regression values were used with shared kernels across layers.
The log transform was performed to optimize the networks for CVD risk stratification. Given that  the total per-patient Agatston scores are non-linearly stratified into CVD risk categories, more importance was assigned to lower Agatston scores -- the log transform induces higher penalties for lower values, thus forcing higher precision for lower calcium burden.
%The results showed good correlation between the scores, but the scores were overestimating calcium burden, visual inspection showed that overestimated image slices contained noise.
%In the first experiment, axial slices cropped to the bounding box around the heart were used as input to the ConvNet.
%In the second experiment, the aim was to increase robustness to noise because in scans with high noise levels, especially those reconstructed with sharp kernels, very small calcifications frequently resemble noise. Thus, Gaussian noise ($\sigma=0.001$) was added to the input slices.
%In the third experiment, the influence of regularization to the network was evaluated by adding a regularization term to the loss function:
%$$
%C(y,\hat{y}) = \left| y - \hat{y} \right| tanh\left(\hat{y}\right) .
%$$
%The rationale behind this is that the network would be more conservative in the prediction of calcium burden in image slices with little calcium, and less restrictive in image slices with a lot of calcium. Adding $tanh(\hat{y})$ as a regularization term allows the predicted values to be as low as possible, while limiting the influence of high target values on the total loss in the mini-batches.

\section{Evaluation}
The total subject Agatston and volume scores obtained automatically by the ConvNet regressor were compared with manually determined scores. Two-way intraclass correlation coefficient (ICC) for absolute agreement with 95\% confidence interval was computed. In addition, for each subject, a CVD risk category was determined based on the Agatston score. Agreement between reference and automatic CVD risk categorization was assessed using accuracy and Cohen's linearly weighted $\kappa$. 

\section{Results and Discussion}

Six scans were excluded from the experiments (3 scans from the training, 2 from the validation and 1 from the test set) because heart localization failed in one scan, and in the other scans the heart was larger than the allowed maximum size of the input. The training set contained 47,007 image slices.
 
Table~\ref{tab:scores} lists ICC values for Agatston and CAC volume scores for each experiment. The results show that the best performance was achieved when weights were not shared between kernels across layers and no transformation of the target values was performed. Furthermore, a slightly better performance was achieved in prediction of Agatston than of CAC volume scores. Given that the Agatston score depends on size (area) and on maximum intensity of calcifications, while the volume score depends on size (number of voxels) only, this is not a surprising result. Namely, ConvNets perform the analysis by determining texture filters. 

\begin{table}[ht!]
	\centering
	\caption[]{ICC showing agreement between automatically computed and reference Agatston and volume scores with 95\% confidence intervals in brackets. Results were obtained for the ConvNet trained with (i) separate kernels, (ii) with shared kernels across layers (\chains), (iii) with separate kernels and a log transform of the target values ($ln(y + 1)$), and (iv) with the shared kernels and a log transform of the target values. The best results are shown in bold.}
	\label{tab:scores}
	\begin{tabular}{C{2cm}C{1.5cm}C{.5cm}|C{3.5cm}C{3.5cm}}
		Experiment  & $ln(y + 1)$ & \chains & Agatston & Volume\\
		\hline
		i   &  &  & \bf{0.98 (0.97--0.98)}  & \bf{0.97 (0.96--0.97)} \\
		ii  &  & \checkmark  & 0.97 (0.97--0.98) & 0.96 (0.95--0.97) \\
		iii & \checkmark  & & 0.96 (0.95--0.97) & 0.95 (0.94--0.96) \\
		iv  & \checkmark & \checkmark & 0.94 (0.93--0.95) & 0.93 (0.92--0.95) \\
	\end{tabular}
\end{table}

Table~\ref{tab:CVDrisk} lists accuracy and Cohen's linearly weighted kappa showing agreement in CVD risk category assignment between manually defined and automatically predicted Agatston scores. These results also show that the best performance was achieved when weights were not shared between kernels across layers. However, unlike in CAC score prediction, better results were achieved when the target Agatston scores were log transformed. As hypothesized, log transformation increased precision and accuracy of low valued Agatston scores and therefore of risk categorization.

\begin{table}[ht!]
	\centering
	\caption[]{Accuracy and linearly weighted $\kappa$ showing agreement between automatic and manual risk  stratification of subjects ($<$1, 1--10, 10--100, 100--400, $\geq$400). The results are shown for experiments when the ConvNet was trained with (i) separate kernels, (ii) with shared kernels across layers (\chains), (iii) with separate kernels and a log transform of the target values ($ln(y + 1)$), and (iv) with the shared kernels and a log transform of the target values. The best results are shown in bold.}
	\label{tab:CVDrisk}
	\begin{tabular}{C{2cm}C{1.5cm}C{.5cm}|C{3.5cm}C{3.5cm}}
		Experiment  & $ln(y + 1)$ & \chains & Accuracy & $\kappa$\\
		\hline
		i   &  &  & 0.80  & 0.85 \\
		ii  &  & \checkmark  & 0.57 & 0.67 \\
		iii & \checkmark  & & \bf{0.85} & \bf{0.90} \\
		iv  & \checkmark & \checkmark & 0.82 & 0.88 \\
	\end{tabular}
\end{table}

\begin{figure}[ht!]
	\centering
	\subfloat{\includegraphics[width=.25\linewidth]{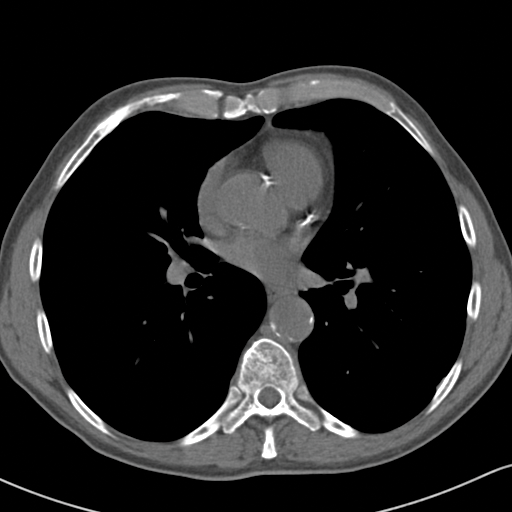}}\quad
	\subfloat{\includegraphics[width=.25\linewidth]{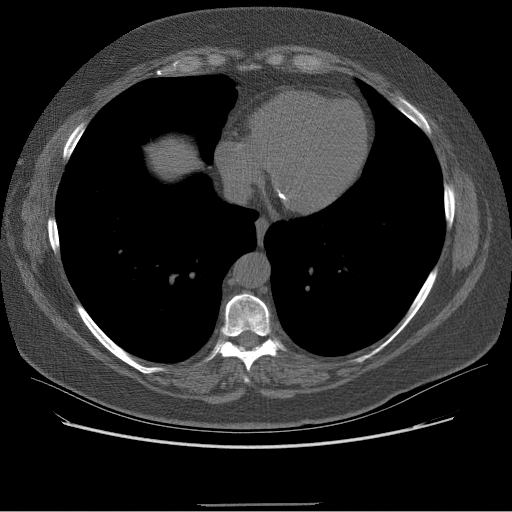}}\quad
	\subfloat{\includegraphics[width=.25\linewidth]{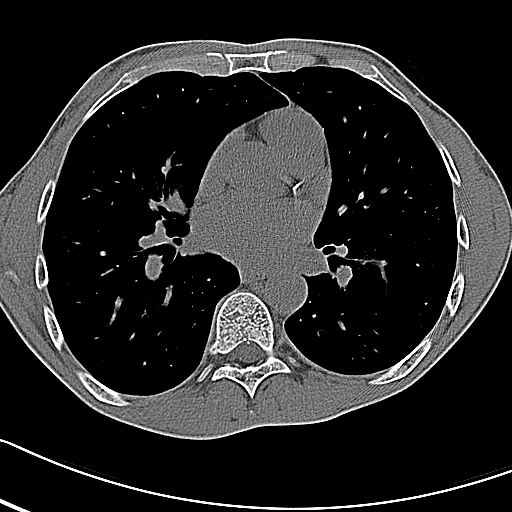}}
	\\
	\subfloat{\includegraphics[width=.25\linewidth]{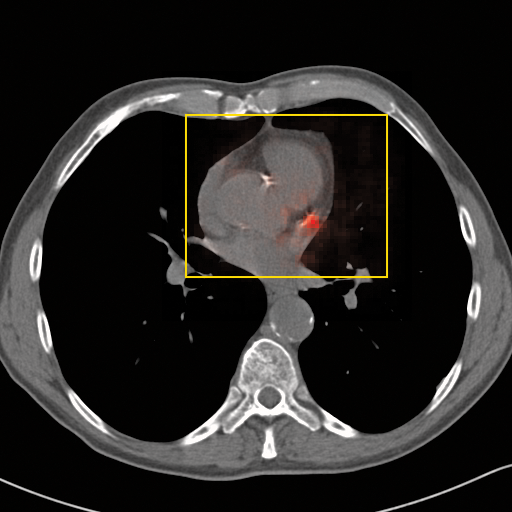}}\quad
	\subfloat{\includegraphics[width=.25\linewidth]{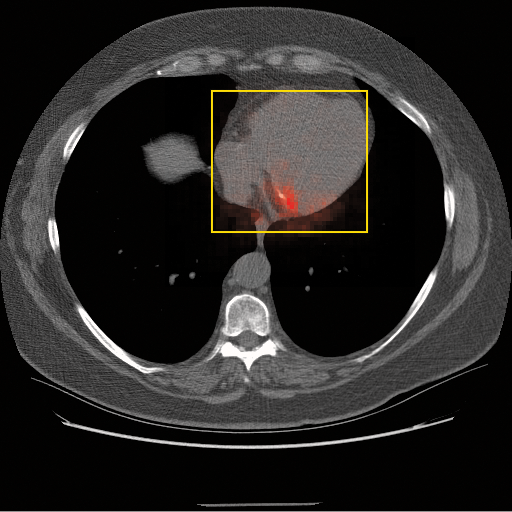}}\quad
	\subfloat{\includegraphics[width=.25\linewidth]{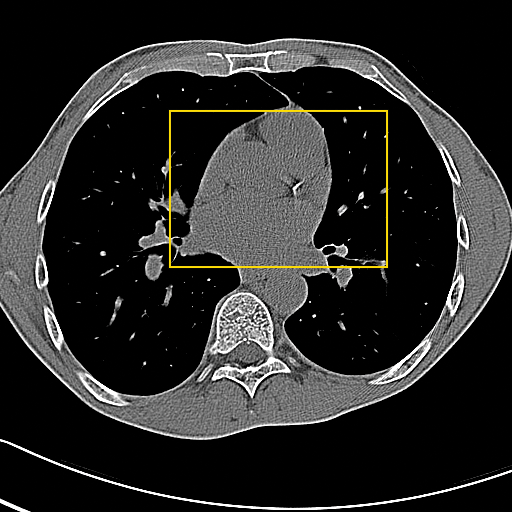}}
	\caption{Top: input image slices. Bottom: automatically determined heart bounding boxes (yellow) and attention of the regression ConvNet (red). Left: correct risk categorization of an image slice with an Agatston score of 62 and a predicted score of 68. Note that the ConvNet focused on the CAC but disregarded calcium in the aorta. Middle: overestimation of CVD risk because of incorrect attention to a calcified lesion that is not in the coronaries. Right: underestimation of risk category caused by missing a calcification in the left coronary artery.}
    \label{fig:examples}
\end{figure}

To illustrate that the proposed method obtains results by detecting CAC, transposed convolutions were performed to obtain feature maps back to the third convolutional layer. A joint response of the network was plotted as a heatmap (Figure~\ref{fig:examples}) to serve as and indication of the area of interest that determined the predicted CAC score.  

Earlier studies performing automatic calcium scoring in cardiac or chest scans first identified calcified lesions and subsequently computed calcium scores. Several methods developed for CAC scoring in cardiac CT scans reported accurate quantification but are typically not directly applicable for calcium scoring in chest CT \cite{Wolt16}. To the best of our knowledge, three methods were developed for CAC scoring in chest CT scans acquired in lung cancer screening settings. The studies evaluated performance of their methods differently and it is therefore not possible to directly compare their performance. Xie et al. \cite{Xie14} analyzed a set of 41 scans from the NLST and reported agreement in the assignment of subjects in four CVD risk categories (0--10, 10--100, 100--400, $\geq$400) of 59\%. Isgum et al. \cite{Isgu12} and Lessmann et al. \cite{Less16} performed calcium scoring in scans from the NELSON trial and reported agreement in assignment of subjects in five CVD risk groups ($<$1, 1--10, 10--100, 100--400, $\geq$400). Both studies reported high accuracy of 82\% and 84\%, respectively. However, both studies analyzed scans that were acquired with a standardized protocol. In this study scans were used from 31 hospitals acquired on 31 different scanners with a variety in protocols. This resulted in very diverse set of chest scans.

Training the networks took approximately 7~hours per network, but testing took less than 2~seconds per scan on a state-of-the-art GPU, including localization of the heart.

The ConvNet predicting calcium score analyzed axial slices cropped to the bounding box around the heart. Although the method to generate these bounding boxes was robust, as determined by visual inspection; in future work, we aim to investigate performance of the method in full axial slices. Furthermore, the method has been evaluated using chest CT scans. In future work we will evaluate application to dedicated cardiac scans and other scans that visualize the heart.
%\textcolor{red}{can you analyze results per scanner/acquisition kernel?}

\section{Conclusion}
A method for direct determination of coronary artery calcium scores from CT images has been presented. Unlike previous methods, this method allows for real-time quantification of coronary calcium burden without the need for identification or segmentation of individual coronary artery calcifications. The results demonstrate that the method is able to assign subjects undergoing lung cancer screening to CVD risk groups with high accuracy. 

\section*{Acknowledgments}
%*******
The authors thank the National Cancer Institute for access to NCI's data collected by the National Lung Screening Trial. The statements contained herein are solely those of the authors and do not represent or imply concurrence or endorsement by NCI.

\end{document}